\documentclass[conference,a4paper]{APSIPA2021}
\usepackage{multirow}
\usepackage{array,graphicx}
\usepackage{float}
\usepackage{amsmath}
\usepackage[psamsfonts]{amssymb}
\usepackage{amsxtra}
\usepackage{threeparttable}
\usepackage{bm} % bold symbol
\usepackage[noend]{algpseudocode}
\usepackage{algorithmicx,algorithm}
\usepackage{float}
\begin{document}

\title{Acceleration of Subspace Learning Machine via Particle
Swarm Optimization and Parallel Processing}

\author{%
\authorblockN{%
Hongyu Fu\authorrefmark{1},  Yijing Yang\authorrefmark{1}, Yuhuai Liu\authorrefmark{1}, 
Joseph Lin\authorrefmark{1}, Ethan Harrison\authorrefmark{3}, Vinod K. Mishra\authorrefmark{2} 
and C.-C. Jay Kuo \authorrefmark{1}
}
\authorblockA{%
\authorrefmark{1}
University of Southern California, Los Angeles, California, USA \\
}
% {
\authorblockA{%
\authorrefmark{2} 
Army Research Laboratory, Aberdeen Proving Ground, Maryland, USA \\
}
\authorblockA{%
\authorrefmark{3}
Mira Costa Highschool, Manhattan Beach, California, USA \\
}
% }
}

\maketitle
\thispagestyle{empty}

\begin{abstract}

Built upon the decision tree (DT) classification and regression idea,
the subspace learning machine (SLM) has been recently proposed to offer
higher performance in general classification and regression tasks. Its
performance improvement is reached at the expense of higher
computational complexity. In this work, we investigate two ways to
accelerate SLM.  First, we adopt the particle swarm optimization (PSO)
algorithm to speed up the search of a discriminant dimension that is
expressed as a linear combination of current dimensions. The search of
optimal weights in the linear combination is computationally heavy. It
is accomplished by probabilistic search in original SLM. The
acceleration of SLM by PSO requires 10-20 times fewer iterations.
Second, we leverage parallel processing in the SLM implementation.
Experimental results show that the accelerated SLM method achieves a
speed up factor of 577 in training time while maintaining comparable
classification/regression performance of original SLM. 

\end{abstract}

\section{Introduction}\label{sec:introduction}

Feature-based classification and regression methods, where feature
extraction and decision-making are two separate modules, have been well
studied for general classification and regression tasks for decades in
the field of machine learning.  In recent years, deep learning (DL)
methods handle feature extraction and decision-making jointly in
training deep neural networks (DNNs) through back propagation (BP).  In
this work, we adopt the traditional machine learning pipeline and aim to
improve the high-performance decision-making (i.e., classification and
regression) module with features as the input. 

Built upon the classical decision tree (DT) \cite{CART} idea, SLM has
been recently proposed in \cite{fu2022subspace}.  SLM offers a
high-performance machine learning solution to general classification and
regression tasks.  Both DT and SLM convert the high-dimensional feature
space splitting problem into a 1D feature space splitting problem. DT
compares each feature in the raw feature space and chooses the one that
meets a certain criterion and searches for the optimal split point, which
is the threshold value. In contrast, SLM conducts a linear combination
of existing features to find a new discriminant 1D space and find the
optimal split point accordingly. It demands to find optimal weights of
the linear combination, which is computationally heavy, in the training
stage. 

It was shown in \cite{fu2022subspace} that SLM achieves better
performance than the support vector machine (SVM) \cite{svm}, DT
\cite{CART}, multilayer perceptron (MLP) \cite{rosenblatt1958perceptron,
ruiyuan} with fewer parameters on several benchmarking machine learning
datasets. The performance improvement is reached at the expense of
higher computational complexity. In this work, we investigate two ideas
to accelerate SLM: 1) proposal of a more efficient weight learning
strategy via particle swarm optimization (PSO)
\cite{kennedy1995particle}, and 2) software implementation optimization
with parallel processing on CPU and GPU. 

For the first idea, the search of optimal weights is accomplished by
probabilistic search in original SLM. No optimization is involved.  The
probabilistic search demands a large number of iterations to achieve
desired results. It is difficult to scale for a high-dimensional feature
space. In contrast, PSO simulates the behavior of a flock of birds,
where each bird is modeled by a particle.  PSO finds the optimal
solution in the search space by updating particles jointly and
iteratively. The acceleration of SLM by PSO requires 10-20 times fewer 
iterations than the original SLM.

For the second idea, SLM's training time can be significantly reduced
through software implementation optimization. In the original SLM
development stage, its implementation primarily relies on public python
packages such as Scikit-learn\cite{sklearn}, NumPy, SciPy \cite{scipy},
etc. After its algorithmic detail is settled, we follow the development
pathway of XGBoost \cite{chen2016xgboost} and LightGBM \cite{lightgbm},
implement a C++ package, and integrate it with Cython. Both multithreaded
CPU and CUDA are supported. 

By combining the above two ideas, accelerated SLM can achieve a speed up
factor of 577 in training as compared to the original SLM while
maintaining comparable (or even better) classification/regression
performance.  The rest of this paper is organized as follows. The SLM
method is reviewed in Sec.  \ref{sec:SLM_review}.  The acceleration of
SLM through PSO is introduced in Sec. \ref{sec:PSO}. The optimization of
software implementation is presented in Sec.
\ref{sec:parallel_process}.  Experimental results are shown in
Sec.~\ref{sec:experiment}. Finally, concluding remarks are given in Sec.
\ref{sec:conclusion}. 

\section{Subspace Learning Machine (SLM)}\label{sec:SLM_review}

\subsection{Methodology}\label{subsec:SLM_overview}

Inspired by the feedforward multilayer perceptron
(FF-MLP)~\cite{ruiyuan}, decision tree (DT)~\cite{CART} and extreme learning machine
(ELM)~\cite{ELM}, a new machine learning model named SLM was proposed by
Fu {\em et al.} in \cite{fu2022subspace} recently.  Its main idea is
subspace partitioning, where an input feature space is partitioned into
multiple discriminant sets in a hierarchical manner.  The whole input
feature space is treated as a parent node and it is partitioned by a
hyperplane into two subsets, where each subset has a lower entropy
value. Each subset is denoted by a child node.  This partitioning
process is conducted recursively until samples in leaf nodes are pure
enough or the sample number is small enough to avoid overfitting.
Majority vote is adopted at each leaf node where all test samples in the
node are predicted as the majority class. Different from traditional DT
methods, SLM allows multiple splitting of a parent node at one level,
which makes the SLM tree wider and shallower to prevent overfitting. 

Instead of partitioning a node based on the original input feature
space, SLM searches for discriminant projections for each parent node
and performs its partitioning in a projected 1D space.  The projected 1D
space from the input feature vector $\bm{x}$ can be expressed as
\begin{equation}\label{eq:gp2}
F_{\bm{a}} = \{f(\bm{a}) \mid f(\bm{a})=\bm{a}^T \bm{x} \},
\end{equation}
where 
\begin{equation}\label{eq:orientation}
\bm{a}=(a_1, \cdots, a_d, \cdots, a_D)^T, \hspace{5mm} || \bm{a} ||=1,
\end{equation}
is the projection vector on the unit hypersphere in $\mathbb{R}^{D}$ 
and $D$ is the dimension of the input feature space. The 1D
subspace can be partitioned into two disjoint sets based on a
threshold, $t_b$, selected by the discriminant feature test (DFT) 
in~\cite{DFT}. They are written as
\begin{eqnarray}
F_{\bm{a},t_b,+} &=&  \{f(\bm{a})  \mid \bm{a}^T \bm{x} \ge t_b \}, \\
F_{\bm{a},t_b,-} &=&  \{f(\bm{a})  \mid \bm{a}^T \bm{x} < t_b  \}.
\end{eqnarray}

We summarize the SLM tree construction as follows.
\begin{enumerate}
\item A discriminant subspace ${S}^{0}$ of dimension ${D}_{0}$ is
selected from the original feature space. 
\item $p$ projection vectors are generated to project feature space
${S}^{0}$ to $p$ 1D subspaces. Among them, the discriminability of each
projected subspace is evaluated using the DFT loss in~\cite{DFT}, to
select the best $q$ discriminant 1D subspaces. Correlation among them
are considered to ensure diversity. 
\item The node is split based on each selected projected subspaces,
denoted as ${S}^{1}$. This completes the node splitting in one level of
the SLM tree. 
\item Conduct node splitting recursively until the stopping criteria are 
met.
\end{enumerate}

\subsection{Probabilistic Selection in SLM}\label{subsec:SLM_proba_selection}

The most important and challenging step in SLM is the selection of
discriminant projection vectors that can separate samples into disjoint
sets of lower entropy. Each projection vector is a linear combination 
of one-hot basis vectors, $\bm{e}_d$, $d=1,\cdots,D$, in form of
\begin{equation}\label{eq:basis}
\bm{a}=a_1 \bm{e}_1 + \cdots, a_d \bm{e}_d + \cdots + a_D \bm{e}_D,
\end{equation}
where $a_d$ represents the coefficient of the corresponding basis vector
$\bm{e}_d$. 

A probabilistic search scheme was proposed in~\cite{fu2022subspace} to
achieve this task. It first orders $\bm{e}_d$ based on its discriminant
power evaluation using the DFT loss from the highest to the lowest.
Then, Eq.~(\ref{eq:basis}) can be rewritten as
\begin{equation}\label{eq:new_basis}
\bm{a}=a'_1 \bm{e'}_1 + \cdots, a'_d \bm{e'}_d + \cdots + 
a'_D \bm{e'}_D.
\end{equation}
where vectors $\bm{e'}_d$ are ordered from the most to the least
discriminant ones.  Then, different sets of coefficients $\{a'_1, a'_2,
\cdots, a'_D\}$ form different projection vectors.  Although random
selection can be performed to find optimal sets of coefficients, the
search space is too large to be practical.  Three hyper-parameters are
proposed to guide the search in probabilistic search as described
below.
\begin{itemize}
\item $P_d$: the probability of selecting coefficient $a'_d$ \\
Since basis vectors are rank ordered in decreasing discriminant power in
Eq.~(\ref{eq:new_basis}), we let $P_d$ decrease exponentially as $d$
increases. In other words, the more discriminant a basis, the higher
likelihood being selected. The resulting $\bm{a}$ tends to be more
discriminant as well. 
\item $A_d$: the dynamic range of coefficient $a'_d$ \\
$a'_d$ is randomly selected among integer values expressed as
\begin{equation}\label{eq:dynamic}
a'_d = 0, \pm 1, \, \pm 2, \, \cdots, \, \pm \lfloor A_d \rfloor,
\end{equation}
where
\begin{equation}\label{eq:exp_a}
A_d= \alpha_0 \exp(-\alpha d).
\end{equation}
The more discriminant a basis, the higher dynamic range in selection.
\item $R$: the number of selected coefficients \\
A subset of $R$ basis vectors out of $D$ are considered in the linear
combination in Eq.~(\ref{eq:new_basis}). The remaining $(D-R)$
corresponding coefficients are set to $0$. This further reduces the
search space for a large $D$ value. 
\end{itemize}

Given a set of selected projection vectors using the probabilistic
search scheme as stated above, the cosine similarity between each pair
is measured.  To ensure the diversity of projected 1D spaces, we prune
projection vectors of higher correlations from the final selected set. 

\section{Acceleration via Adaptive Particle Swarm Optimization (APSO)}\label{sec:PSO}

As described in Sec. \ref{sec:SLM_review}, the probabilistic search is
used in the original SLM to obtain good projection vectors. As the
feature dimension increases, the search space grows exponentially and
the probabilistic search scheme does not scale well. In this section, we
adopt a metaheuristic algorithm called the particle swarm optimization
(PSO) \cite{eberhart2001swarm} to address the same problem with higher
efficiency. 

\subsection{Introduction to PSO}\label{subsec:PSO_introduction}

PSO is an evolutionary algorithm proposed by Kennedy and Eberhart in
1995 \cite{kennedy1995particle}. It finds the optimal solution in the
search space by simulating the behavior of flock using a swarm of
particles. Each particle in the swarm has a position vector and a
velocity vector. Mathematically, the $i$th particle has the following
position and velocity vectors:

\begin{eqnarray}\label{eq:pso_a}
{\bf x}_{i} &=& \left [ x_{1}, x_{2}, \cdots, x_{D} \right ]^T \\
{\bf v}_{i} &=& \left [ v_{1}, v_{2}, \cdots, v_{D} \right ]^T 
\end{eqnarray}

where $D$ is the dimension of the search space, respectively.  The
velocity and position vectors of the $i$th particle are updated in
each iteration via
\begin{eqnarray}
{\bf v}_{i,t+1}=\omega {\bf v}_{i,t} + c_1 r_1 [{\bf pB}_{i,t} - {\bf x}_{i,t}]
+ c_2 r_2 [{\bf gB} - {\bf x}_{i,t}], \label{eq:pso-1} \\
{\bf x}_{i,t+1}={\bf x}_{i,t}+{\bf v}_{i,t}. \label{eq:pso-2}
\end{eqnarray}
Eq. (\ref{eq:pso-1}) has the following parameters:
\begin{itemize}
\item inertia weight $\omega$ used to control the influence of the
current speed of the particle on the update. 
\item ${\bf gB}$: the position vector of the optimal particle in
the population. 
\item ${\bf pB}_{i,t}$: the best position of the $i$th particle in
the history update process up to iteration $t$. 
\item $c_{1}$ and $c_{2}$: two learning factors.
\item $r_{1}$ and $r_{2}$: two uniformly distributed random values
between 0 and 1 used to simulate the stochastic behavior of particles. 
\end{itemize}

The initial position and velocity vectors of all particles are randomly
selected within a preset range. At each iteration, the velocity and
position of each particle are updated according to Eqs.
(\ref{eq:pso-1}) and (\ref{eq:pso-2}), respectively. Furthermore, we
calculate the current loss and update ${\bf gB}$ and ${\bf pB}$
accordingly. If the algorithm converges, the position vector of
the global optimal particle is the output.  The pseudo-codes of the 
PSO are provided in Algorithm 1. 

%%%%%%%%%%%%%%%%%%%%%%%%%%%%%%%%%%%%%%%%%%%%%%%%%%%
\begin{algorithm}[t]
\caption{Particle Swarm Optimization (PSO)}
\hspace*{0.02in} {\bf Input:}
Population, Dimension and MaxIteration\\
\hspace*{0.02in} {\bf Output:}
Best Particle's position vector and loss
\begin{algorithmic}[1]
\State Initialize all particles
\For{$iteration < MaxIteration$}
    \State...
    \For{$i < Population$}
        \State Update $V_{i}$ and $X_{i}$
        \If{$loss_{i} < pbest_{i}$}
            \State Update $pbest_{i}$
            \If{$loss_{i} < gbest$}
                \State Update $gbest$
            \EndIf
        \Else
            \State pass
        \EndIf
        \State i = i+1 //next particle
    \EndFor
    \State iteration = iteration + 1 //next iteration
\EndFor
\State \Return $gbest$
\end{algorithmic}
\end{algorithm}
%%%%%%%%%%%%%%%%%%%%%%%%%%%%%%%%%%%%%%%%%%%%%%%%%%%

\subsection{Adaptive PSO (APSO) in SLM}\label{subsec:APSO}

The PSO algorithm has a couple of variants
\cite{krohling2006coevolutionary, liang2006comprehensive, shi2001fuzzy}.
The search space of the projection vector in our problem is complex and
there are quite a few local minima. A standard PSO can easily get stuck
to the local minima. One variant of PSO, called the adaptive particle
swarm optimization (APSO)\cite{4812104}, often yields better performance
for multimodal problems so that it is adopted as an alternative choice
in the SLM implementation. 

Before an iteration, APSO calculates the distribution of particles in
the space and identify it as one of the following four states:
Exploration, Exploitation, Convergence, and Jumpout. APSO chooses
parameters $c_1$ and $c_2$ in different states dynamically so as to make
particles more effective under the current distribution. They are
summarized below. 
\begin{itemize}
\item In the exploration state, APSO increases the $c_1$ value and
decreases the $c_2$ value.  This allows particles to explore in the whole
search space more freely and individually to achieve better individual
optimal position. 
\item In exploitation state, APSO increases the $c_1$ value slightly and 
decreases the $c_2$ value. As a result, particles can leverage the local
information more to find the global optimum. 
\item In the convergence state, APSO increases both $c_1$ and $c_2$
slightly.  An elite learning strategy is also introduced to promote the
convergence of globally optimal particles, which makes the search more
efficient. 
\item In the jumping-out state, APSO decreases the $c_1$ value and
increases the $c_2$ value.  This allows particles move away from their
current cluster more easily. Thus, the particle swarm can jump out of
the current local optimal solution. 
\end{itemize}

For classification tasks in SLM, particle's position vector is the
projection vector and the DFT loss is utilized as the loss function.  In
original SLM, we split samples into several child nodes at each tree
level.  Since APSO can find the global optimum with the lowest loss
function, only one globally optimal projection is selected per node in
the accelerated version.  As a result, the resulting SLM tree is a
binary tree. Furthermore, in the accelerated SLM algorithm, the top $n$
discriminant features in each node are re-computed and they serve as the input
dimension of the APSO algorithm. We will show in the experimental
section that APSO demands fewer iterations to find the global optimum
than probabilistic search. Thus, the overall search speed can be
improved. The pseudo-codes of the SLM tree building based on APSO is 
given in Algorithm 2. 

%%%%%%%%%%%%%%%%%%%%%%%%%%%%%%%%%%%%%%%%%%%%%%%%%%%
\begin{algorithm}[t]
\caption{SLM Tree Building with APSO}
\begin{algorithmic}[1]
\If{Node meet the split condition}
    \State Run DFT for this node
    \State Select top $n$ channels from DFT result
    \State Run Adaptive PSO with $n$ Dimension
    \State Use $gbest$ vector for partitioning
    \State Get two child nodes for next building process
\Else
    \State Mark this node as a leaf node
\EndIf
\State \Return 
\end{algorithmic}
\end{algorithm}
%%%%%%%%%%%%%%%%%%%%%%%%%%%%%%%%%%%%%%%%%%%%%%%%%%%

\section{Acceleration via Parallel Processing}\label{sec:parallel_process}

The SLM algorithm in \cite{fu2022subspace} was implemented in pure
Python.  While Python code is easy to write, the running speed is
slow. Cython is a superset of Python that bridges the gap between Python
and C. It allows static typing and compilation from Cython to C.
Furthermore, there are a few commonly used math libraries in Python such
as Scipy and scikit-learn. 

The first step towards run-time acceleration is to convert the
computational bottleneck of the SLM implementation from Python to
Cython. The bottleneck is the search of the optimal projection vector.
For a projection vector at a node, we calculate the loss at each split
point so as to find the minimum loss in the 1D space. Since these
computations do not depend on each other, they can be computed in
parallel in principle. Python does not support multithreading due to
the presence of the Global Interpreter Lock (GIL). To this end,
multi-processing was deemed unsuitable due to higher overhead in the
current context. 

Instead, we implement multithreaded C++ extension for the optimal
projection vector search, and integrate it into the remaining Cython
code. In the C++ extension, we spawn multiple threads, where each
thread covers the computation of a fixed number of splits per selected
projection vector. The launched thread number is equal to the minimum of
the physical core number to prevent excessive context switching.  A task
is defined as calculating the loss on a single split of a projected 1D
subspace. These tasks are evenly allocated to each thread wherein each
thread processes them sequentially. 

Compared to CPUs, modern GPUs have thousands of cores and context switching
occurs significantly less frequently. Thus, for GPU processing, each
thread is responsible for a single task in our design. Once the task is
completed, the GPU core moves onto the next available task, if any.
Once the computation is completed, the loss values are returned to the
main process where the minimum loss projection and split are computed. 

%%%%%%%%%%%%%%%%%%%%%%%%%%%%%%%%%%%%%%%%%%%%%%%%%%%
\begin{table*}[t]
\begin{center}
\caption{Comparison of training run-time (in seconds) of 6 settings for 9 
classification datasets.}\label{table:1}
\begin{tabular}{|c|c|c|c|c|c|c|c|c|c|c|}\hline
& Settings & \begin{tabular}[c]{@{}l@{}}circle-and-\\ ring\end{tabular} & \begin{tabular}[c]{@{}l@{}}2-new-\\ moons\end{tabular} & \begin{tabular}[c]{@{}l@{}}4-new-\\ moons\end{tabular} & Iris  & Wine         & B.C.W          & Pima           & Ionoshpere     & Banknote \\
\hline
\parbox[t]{2mm}{\multirow{6}{*}{\rotatebox[origin=c]{90}{SLM}}} 
& A & 7.026&	4.015&	6.019&	82.939&	65.652&	234.597 & 339.077&	248.410&	118.000  \\
& B & 0.287 & 0.188 & \textbf{0.200} & 0.465 & 0.808 & 2.931 & 3.920 & 5.217 & 0.697 \\
& C & 0.346 & 0.0943 & 0.236 & 0.295 & \textbf{0.700} & \textbf{2.657} & 3.263 & 3.838 & 0.439\\
& D & 4.071 & 3.159 & 3.245 & 3.205 & 3.572 & 3.956 & 4.660 & 5.021 & 4.370 \\
& E & 7.597 & 3.146 & 7.200 & 4.188 & 3.106 & 11.605 & 13.892 & 8.123 & 8.588 \\
& F & \textbf{0.270} & \textbf{0.0887} & 0.278 & \textbf{0.192} & 0.831 & 2.837 & \textbf{2.738} & \textbf{1.371} & \textbf{0.619}
\\ \hline \hline
\parbox[t]{2mm}{\multirow{6}{*}{\rotatebox[origin=c]{90}{SLM Forest}}} 
& A & 160.264&	107.695	&157.307&	2333.514&	2605.288 &5067.356&	10024.238&	6582.104&	3020.587 \\
& B & 8.507 & 5.272 & 8.471 & 17.142 & 26.254 & 134.301 & 101.631 & 152.625 & 40.365 \\
& C & 7.910 & 4.714 & 8.170 & 13.068 & 22.176 & 116.843 & 88.972 & 137.390 & 16.341 \\
& D & \textbf{7.708} & 4.954 & \textbf{7.678} & 5.946 & \textbf{15.066} & 98.969 & 48.987 & 80.547 & 16.511 \\
& E & 220.021 & 96.712 & 219.002 & 105.768 & 94.788 & 360.089 & 396.989 & 247.069 & 259.000 \\
& F & 8.75	&\textbf{4.058}&	9.02&	\textbf{5.822}&	20.081&	\textbf{98.490}&	\textbf{31.684}&	\textbf{75.964}&	\textbf{14.054}
\\ \hline \hline
\parbox[t]{2mm}{\multirow{6}{*}{\rotatebox[origin=c]{90}{SLM Boost}}} 
& A & 143.332 &	116.156&	151.904&	2347.904&	2591.919&	7118.338	& 9887.949 &	6395.744&	3412.654	\\
& B & 7.459&	5.185&	\textbf{7.533}&	16.087&	21.873&	117.01&	104.47&	148.586&	23.443 \\
& C & 7.987&	\textbf{3.434}&	8.293&	12.144&	15.624&	115.226&	80.273&	131.712&	\textbf{14.207} \\
& D & 7.741&	4.894&	8.182&	6.093&	\textbf{15.573}&	97.599&	46.057&	\textbf{76.589}& 15.588 \\
& E & 233.071&	97.541&	233.347&	107.709&	95.534&	362.295&	396.989&	247.069&	262.546 \\
& F & \textbf{6.837}&	3.581&	8.425&	\textbf{5.915}&	20.159&	\textbf{94.413}&	\textbf{40.413}&	79.012&	15.516  \\ \hline
\end{tabular}
\end{center}
\end{table*}
%%%%%%%%%%%%%%%%%%%%%%%%%%%%%%%%%%%%%%%%%%%%%%%%%%%

%%%%%%%%%%%%%%%%%%%%%%%%%%%%%%%%%%%%%%%%%%%%%%%%%%%
\begin{table*}[t]
\begin{center}
\caption{Comparison of training run-time (in seconds) of 6 settings for 6 
regression datasets.}\label{table:2}
\begin{tabular}{|c|c|c|c|c|c|c|c|}\hline
& Settings & Make Friedman1 & Make Friedman2 & Make Friedman3 & Boston & California\_housing & Diabetes \\
\hline
\parbox[t]{2mm}{\multirow{6}{*}{\rotatebox[origin=c]{90}{SLR}}} 
& A & 152.291&	60.232&	186.535&	153.173&	756.868&	147.933  \\
& B & 2.136&	0.653&	2.137&	1.255&	26.417&	1.136 \\
& C & 0.937&	\textbf{0.331}&	2.100&	0.780&	5.625&	0.608 \\
& D & 1.577&	0.448&	3.832&	1.31&	\textbf{5.508}&	0.722 \\
& E & 29.742&	28.294&	44.511&	30.608&	96.319&	27.162 \\
& F &\textbf{0.639}&	0.332&	\textbf{0.755}&	\textbf{0.675}&	5.891&	\textbf{0.322} \\ \hline \hline
\parbox[t]{2mm}{\multirow{6}{*}{\rotatebox[origin=c]{90}{SLR Forest}}} 
& A & 5406.271&	3346.595&	7459.372&	6454.979&	21092.217&	4413.089  \\
& B & 62.527&	19.244&57.117&	55.204&	1049.859&	46.622 \\
& C & 35.576&	13.533&	63.551&	30.943&	254.415	& 29.606\\
& D & 19.408&	\textbf{6.48}&	42.182&	18.04&	\textbf{125.104}&	17.242 \\
& E & 1025.801&	940.359&	1051.498&	792.33&	3190.336&	775.153 \\
& F & \textbf{17.993}&13.718&\textbf{19.898}&\textbf{11.178}&193.741&\textbf{10.879}\\ \hline \hline
\parbox[t]{2mm}{\multirow{6}{*}{\rotatebox[origin=c]{90}{SLR Boost}}} 
& A & 4678.86&	1955.517&	6770.808&	4097.66	& 21612.672&	4200.828 \\
& B & 62.156&	19.724&	59.435&	38.154&	817.688&	34.767 \\
& C & 26.704&	10.546&	70.716&	29.149&	159.434	& 20.528 \\
& D & 19.293&	\textbf{6.551}&	51.222&	19.817&	\textbf{124.702}&	38.897 \\
& E & 919.874&	920.619&	1834.926&	910.271&	3336.018&	901.092 \\
& F & \textbf{14.170}&	13.114&	\textbf{24.759}&\textbf{12.186}&195.459&\textbf{10.594} \\ \hline
\end{tabular}
\end{center}
\end{table*}
%%%%%%%%%%%%%%%%%%%%%%%%%%%%%%%%%%%%%%%%%%%%%%%%%%%

\section{Experiments} \label{sec:experiment}

{\bf Datasets.} We follow the experimental setup in \cite{fu2022subspace}
and conduct performance benchmarking on 9 classification datasets (i.e.,
Circle-and-ring, 2-new-moons, 4-new-moons, Iris, Wine, B.C.W, Pima,
Ionoshpere, and Banknote) and 6 regression datasets (i.e., Make
Friedman1, Makefriedman2, Makefriedman3, Boston, California housing and
Diabetes). For details of these 15 datasets, we refer readers to
\cite{fu2022subspace}. 

{\bf Benchmarking Algorithms and Accelation Settings.} Ensemble methods are commonly used in the machine learning field to
boost the performance. An ensemble model aims to obtain better
performance than each constituent model alone.  Inspired by the random
forest (RF) \cite{RF} and XGBoost for DT, SLM Forest and SLM Boost are
bagging and boosting methods for SLM, respectively. The counter part of
SLM for the regression task is called the subspace learning regressor
(SLR). Again, SLR has two ensemble methods; namely, SLR Forest and SLR
Boost. They are detailed in \cite{fu2022subspace}. 

We compare the run-time of six settings of SLM, SLM Forest and SLM Boost
three methods in Table \ref{table:1} and six settings of SLR, SLR
Forest and SLR Boost in Table \ref{table:2}. They are:
\begin{itemize}
\item[A] Probabilistic search in pure Python, 
\item[B] Probabilistic search in Cython and C++ singlethreaded, 
\item[C] Probabilistic search in Cython and C++ multithreaded, 
\item[D] Probabilistic search in CUDA/GPU,
\item[E] APSO in pure Python,
\item[F] APSO in Cython and C++ multithreaded, 
\end{itemize}
For a given dataset, settings [A]-[D] provide the same classification
(or regression) performance in terms of classification accuracy (or
regression MSE) while settings [E] and [F] offer the same classification
accuracy (or regression MSE). Their main difference lies in the model
training time.  All settings are implemented on Intel(R) Xeon(R) CPU
E5-2620 v3 @2.40GHz with 12 cores 24 threads and Nvidia Quadro M6000
GPU. All hyperparameters are the same.  The number of trees for two
SLM/SLR ensemble methods (i.e., SLM/SLR Forest and SLM/SLR Boost) is set
to 30. 

{\bf Accelation by Parallel Processing.} We compare the run-time of SLM,
SLM Forest and SLM Boost for the 9 classification datasets in Table
\ref{table:1} and that of SLR, SLR Forest and SLR Boost for the 6
regression datasets in Table \ref{table:2}.  The shortest run-time is
highlighted in bold face. We see from the tables that simply changing
the implementation from Python to C++ yields a speed-up factor of x40 to
x100 across all datasets.  After that, the speed-up is incremental.
Multithreading should lead to increased performance in theory. Yet, it
is greatly influenced by the state of the machine in practice. For
example, caching, task scheduling, and whether other processors are used
by other processes.  Moreover, additional overhead occurs in spawning,
managing, and destroying threads. Typically, a thread of longer
computation is more suitable for multithreading.  Thus, the performance
improvement in smaller datasets, where computations take less time, is
less pronounced.  Similarly, for a single SLM tree training, the model
training time on GPU is notably worse due to the overhead of moving data
to GPU memory. As the training scales (e.g., with ensembles of SLM/SLR trees
such as SLM/SLR Forest and SLM/SLR Boost), CUDA/GPU generally outperforms
multithreading. 

%%%%%%%%%%%%%%%%%%%%%%%%%%%%%%%%%%%%%%%%%%%%%%%%%%%
%Table 3
\begin{table*}[t]
\begin{center}
\caption{Comparison of classification accuracy (in terms of \%) of SLM,
SLM Forest and SLM Boost probabilistic search and APSO acceleration
methods on 9 classification datasets.}\label{table:3}
\begin{tabular}{|l|l|l|l|l|l|l|l|l|l||l|} \hline
Settings       & \begin{tabular}[c]{@{}l@{}}circle-and-\\ ring\end{tabular} & \begin{tabular}[c]{@{}l@{}}2-new-\\ moons\end{tabular} & \begin{tabular}[c]{@{}l@{}}4-new-\\ moons\end{tabular} & Iris  & Wine         & B.C.W          & Pima           & Ionoshpere     & Banknote       & Number of Iteration \\ \hline
SLM [A,B,C,D]  & 96.50                                                      & 100                                                    & 99.75                                                  & 98.33 & 98.61        & 97.23          & 77.71          & 90.07          & 99.09          & 1000                \\
SLM [E,F]               & 96.50                                                      & 100                                                    & 99.75                                                  & 98.33 & \textbf{100} & \textbf{98.68} & \textbf{77.71} & \textbf{92.19} & \textbf{99.64} & \textbf{110}   \\ \hline \hline 
SLM Forest [A,B,C,D] & 100                                                        & 100                                                    & 100                                                    & 98.33 & \textbf{100} & 97.36          & \textbf{79.00} & \textbf{95.71} & 99.81          & 1000                \\
SLM Forest [E,F]         & 100                                                        & 100                                                    & 100                                                    & 98.33 & 98.61        & \textbf{97.80} & 77.71          & 95.03          & \textbf{100}   & 110 \\ \hline \hline
SLM Boost [A,B,C,D]  & 100                                                        & 100                                                    & 100                                                    & 98.33 & 100          & \textbf{98.83} & 77.71          & 94.33          & 100            & 1000                \\
SLM Boost [E,F]          & 100                                                        & 100                                                    & 100                                                    & 98.33 & 100          & 97.80          & \textbf{78.34} & 94.33          & 100   & \textbf{110} \\ \hline %\hline
\end{tabular}
\end{center}
\end{table*}
%%%%%%%%%%%%%%%%%%%%%%%%%%%%%%%%%%%%%%%%%%%%%%%%%%%

%%%%%%%%%%%%%%%%%%%%%%%%%%%%%%%%%%%%%%%%%%%%%%%%%%%
%Table 4
\begin{table*}[t]
\begin{center}
\caption{Comparison of regression mean-squared-errors (MSE) of SLR, SLR
Forest and SLR Boost probabilistic search and APSO acceleration methods
on 6 regression datasets.}\label{table:4}
\begin{tabular}{|l|l|l|l|l|l|l||l|}\hline
settings   & Make Friedman1 & Makefriedman2  & Makefriedman3  & Boston         & California housing & Diabetes        & Number of Iteration \\ \hline
SLR [A,B,C,D]  & 20.46          & 125830         & 0.078          & 65.68          & 1.172              & 5238.8          & 2000           \\ 
SLR [E,F] & \textbf{13.52} & \textbf{12001} & \textbf{0.022} & \textbf{27.88} & \textbf{0.575}     & \textbf{4018.7} & \textbf{110}   \\ \hline\hline
SLR Forest [A,B,C,D]& 5.46           & \textbf{1641}  & \textbf{0.007} & \textbf{13.18} & \textbf{0.407}     & \textbf{2624.4} & 2000                \\ 
SLR Forest [E,F]    & \textbf{4.68}  & 2144           & 0.012          & 13.48          & 0.419              & 2640.7       & \textbf{110}\\ \hline\hline
SLR Boost [A,B,C,D] & \textbf{3.85}  & 670            & 0.008          & 13.07          & 0.365              & \textbf{2710.0} & 2000                \\ 
SLR Boost [E,F]  & 3.95           & \textbf{662}   & \textbf{0.005} & \textbf{12.91} & \textbf{0.341}     & 2711.2   & \textbf{110} \\ \hline%\hline
\end{tabular}
\end{center}
\end{table*}
%%%%%%%%%%%%%%%%%%%%%%%%%%%%%%%%%%%%%%%%%%%%%%%%%%%

{\bf Acceleration by APSO.} To demonstrate the effectiveness of APSO, we
compare the classification and regression performance of probabilistic
search and APSO in Tables \ref{table:3} and \ref{table:4},
respectively.  Their performance in terms of classification accuracy and
regression MSE is comparable.  As shown in the last column of both
tables, the iteration number of APSO is about 10\% (for SLM) and 5\% (for
SLR) of that of probabilistic search. 

{\bf Joint Acceleration by Parallel Processing and APSO.} We can see the
effect of joint acceleration of parallel processing and APSO from Tables
\ref{table:1} and \ref{table:2} method F.  We observe that the single-threaded
C++ implementation of APSO has a comparable speed of the GPU version of
probabilistic search. We have not yet done the GPU version of APSO
acceleration but expect that it will provide another level of boosting
in shortening the training run time. 

In summary, APSO acceleration reduces the time complexity of SLM/SLR by
reducing the number of iterations to 5-10\%. Its C++ implementation and
parallel process reduces the training time of each iteration by a factor
of x40 to x100. With the combination of both, APSO SLM/SLR with C++
implementation gives the best overall performance in training time,
classification accuracy and regression error. As compared to the
original SLM/SLR, its training time can be accelerated by a factor
up to 577.

\section{Conclusion and Future Work} \label{sec:conclusion}

Two ways were proposed to accelerate SLM in this work.  First, we
applied the particle swarm optimization (PSO) algorithm to speed up the
search of a discriminant dimension, which was accomplished by
probabilistic search in original SLM. The acceleration of SLM by PSO
requires 10-20 times fewer iterations.  Second, we leveraged parallel
processing in the implementation. It was shown by experimental results
that, as compared to original SLM/SLR, accelerated SLM/SLR can achieve a
speed up factor of 577 in training time while maintaining comparable
classification and regression performance. 

The datasets considered in \cite{fu2022subspace} and this work are of
lower dimensions. Recently, research on unsupervised representation of
images and videos has been intensively studied, e.g.,
\cite{chen2020pixelhop, yang2022design, kuo2019interpretable,
zhou2021uhp, zhou2021unsupervised,zhou2022gusot, zhu2022pixelhop}, learned
representations and their labels can be fed into a classifier/regressor
for supervised classification and regression.  This leads to the
traditional pattern recognition paradigm, where feature extraction and
classification modules are cascaded for various tasks.  The feature
dimensions for these problems are in the order of hundreds. It is
important to develop SLM for high dimensional image and video data to
explore its full potential.

\bibliographystyle{ieeetr}
\bibliography{refs.bib}

% \begin{thebibliography}{1}
% \bibitem{1}
% \end{thebibliography}

\end{document}